\documentclass[10pt,twocolumn,letterpaper]{article}

\usepackage{iccv}
\usepackage{times}
\usepackage{epsfig}
\usepackage{graphicx}
\usepackage{amsmath}
\usepackage{amssymb}
\usepackage{caption}
\usepackage{subcaption}
\usepackage{multirow}
\usepackage{tabularx}
\usepackage{mathtools}
\usepackage{lscape}
\usepackage{array}
\usepackage{booktabs}
\usepackage{graphicx}
\usepackage{xcolor}
\usepackage[accsupp]{axessibility}

\usepackage[breaklinks=true,bookmarks=false]{hyperref}

\iccvfinalcopy 

\def\iccvPaperID{9823} 

\ificcvfinal\fi

\begin{document}

\title{Just One Moment:\\Structural Vulnerability of Deep Action Recognition against One Frame Attack}

\author{Jaehui Hwang, Jun-Hyuk Kim, Jun-Ho Choi, and Jong-Seok Lee\\
School of Integrated Technology, Yonsei University, Korea\\
{\tt\small \{jaehui.hwang, junhyuk.kim, idearibosome, jong-seok.lee\}@yonsei.ac.kr}
}

\maketitle
\ificcvfinal\thispagestyle{empty}\fi

\begin{abstract}
   The video-based action recognition task has been extensively studied in recent years.
    In this paper, we study the structural vulnerability of deep learning-based action recognition models against the adversarial attack using the one frame attack that adds an inconspicuous perturbation to only a single frame of a given video clip.
    Our analysis shows that the models are highly vulnerable against the one frame attack due to their structural properties.
    Experiments demonstrate high fooling rates and inconspicuous characteristics of the attack.
    Furthermore, we show that strong universal one frame perturbations can be obtained under various scenarios.
    Our work raises the serious issue of adversarial vulnerability of the state-of-the-art action recognition models in various perspectives.
\end{abstract}

\section{Introduction}
\label{sec: introduction}

Human action recognition using videos has been extensively studied in recent years thanks to the development of deep network-based algorithms based on the abundance of computational resources and data~\cite{herath2017going}.
From a network design perspective, the distinguished main research topic of action recognition is how to model temporal information residing in video clips.
Concerning this, various attempts have been made, such as utilizing the long short-term memory (LSTM) module~\cite{donahue2015long} or the optical flow~\cite{carreira2017quo}, but recently, 3D convolutional neural network (CNN)-based action recognition models are widely used.
To improve the performance and efficiency of 3D CNN-based action recognition models, various mechanisms in the temporal dimension have been proposed, such as frame selection \cite{feichtenhofer2019slowfast} and convolutional operations \cite{carreira2017quo,wang2018non,tran2019video}.

Many researchers have found the vulnerability of deep learning-based algorithms against so-called adversarial attacks, which add an inconspicuous perturbation to input data to mislead a target model to produce wrong output.
It has been reported that many state-of-the-art deep image classification methods are highly vulnerable to the adversarial attacks~\cite{su2018robustness} and can raise severe security concerns~\cite{goodfellow2014explaining}.
On the other hand, there are not many studies on the vulnerability of video-based deep action recognition systems.

\begin{figure}
	\centering
	\includegraphics[width=1.0\columnwidth]{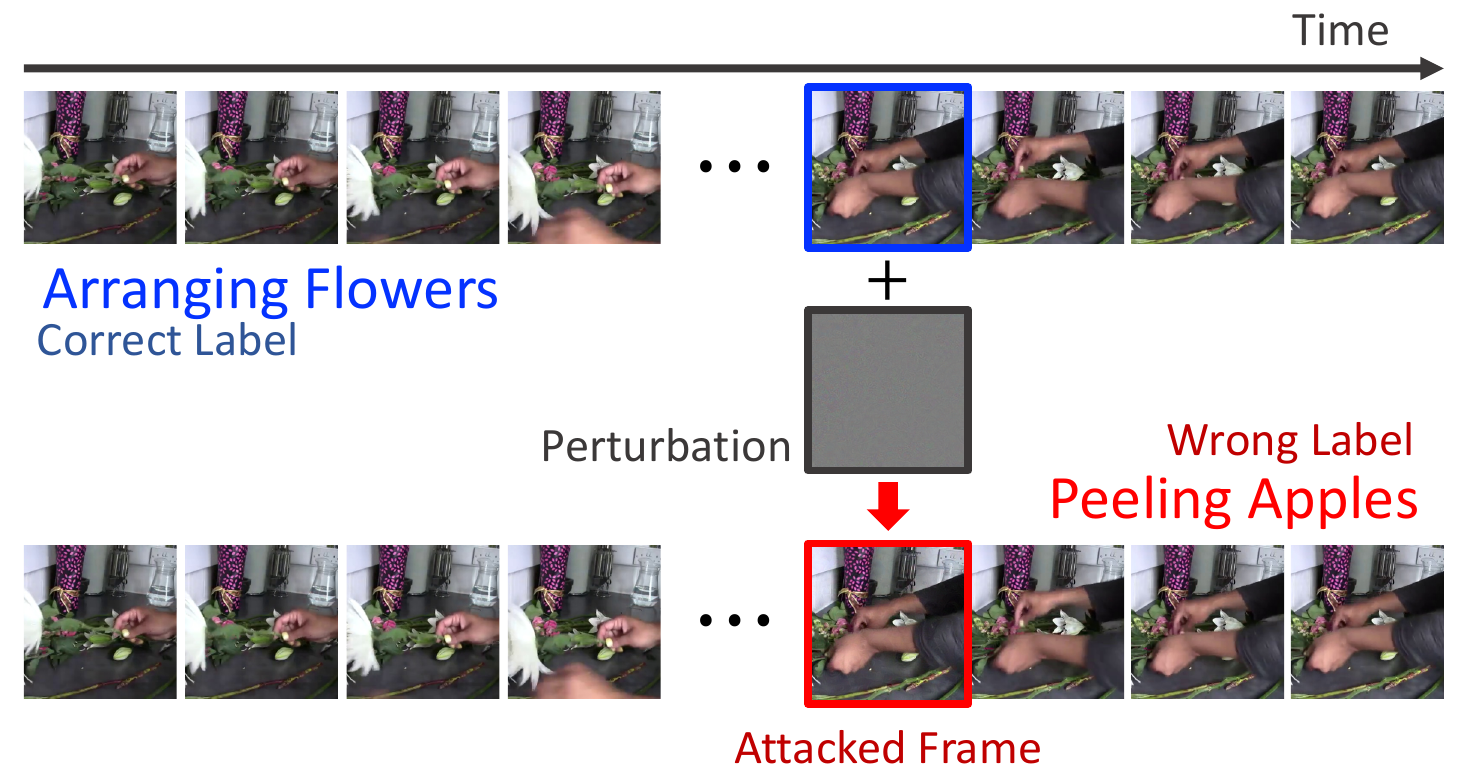}
	\caption{Overall illustration of one frame attack targeting a vulnerable frame. With only one attacked frame, the target action recognition model wrongly classifies a given video in Kinetics-400 \cite{kay2017kinetics} as ``peeling apples'' instead of ``arranging flowers.''}
	\label{fig:teaser}
\end{figure}

We argue that the ways of modeling temporal information in deep models for action recognition have significant impact on the vulnerability of the models, which we denote as \emph{structural vulnerability}.
Although there exist a few attempts for adversarial attacks on action recognition systems~\cite{wei2019sparse,naeh2020patternless}, they do not sufficiently consider the structural vulnerability of recent action recognition models.
As a result, they do not fully satisfy two criteria that a successful adversarial attack should meet: 1) achieving a high fooling rate and 2) keeping adversarial perturbation invisible to conceal that the video clip is attacked.
Wei \etal \cite{wei2019sparse} proposed an attack method of perturbing only a few frames of a video clip, in order to reduce computational resources and achieve inconspicuousness. 
However, this method targets only LSTM-based models and relies on the particular property of LSTM, i.e., temporal propagation of information.
Thus, it does not achieve a high fooling rate on the latest CNN-based action recognition models (see Table \ref{table:performs}).
On the other hand, Pony \etal \cite{naeh2020patternless} proposed an attack method to find a sequence of flickering perturbation that changes the overall color of a given video clip over time.
However, it is known that artifacts changing in the temporal dimension are more detectable than spatial artifacts by human observers \cite{ninassi2009considering,zhao2010evaluating}.
Therefore, this adversarial attack does not sufficiently satisfy the second criterion, inconspicuousness (see Figure \ref{fig:subjective test}). 

In this paper, we discover the structural vulnerability of recent CNN-based deep action recognition models, which has not been explored previously to the best of our knowledge.
Using this vulnerability, we also show that perturbation in just a single vulnerable frame of a video clip can significantly degrade the accuracy of deep action recognition models, as illustrated in \figurename~\ref{fig:teaser}.
The attacked frame is shown only for 33 or 40 milliseconds when the target video clip has 30 or 25 frames per second (FPS), which is hardly perceivable to human observers.
The main contributions of this work can be summarized as follows.

\begin{itemize}
	\item
	We investigate the vulnerability caused by the structural property of deep models using three state-of-the-art CNN-based deep action recognition models and examine what factors of these models make them highly vulnerable against adversarial attacks.
	We show that the efforts to efficiently model temporal information induce the vulnerability issue.
	\item
	We show the possibility of so-called \emph{one frame attack} on action recognition models in a white-box attack scenario.
	When only one vulnerable frame found by our analysis is perturbed with a gradient-based adversarial attack method, this perturbation can easily defeat deep learning-based action recognition systems.
	This one frame attack can fool the state-of-the-art video-based action recognition models with fooling rates of almost 100\%.
	In addition, this adversarial attack is inconspicuous, which is demonstrated via a subjective experiment. 
    \item
	We further explore video-agnostic universal perturbation based on the one frame attack.
	We show that the universal perturbation, which is found from a small number of videos, can affect other input video clips with high fooling rates.
	Besides, the one frame attack can be effectively applied to time-invariant scenarios where the perturbation is added to the input video clip with an unknown temporal offset.
	
\end{itemize}

\section{Related work}
\label{sec:related_work}
\subsection{Action recognition}
Recently, the performance of action recognition has been significantly improved, along with the development of deep neural networks~\cite{carreira2017quo,donahue2015long,feichtenhofer2019slowfast,ji20123d,simonyan2014two,tran2015learning,tran2019video,wang2018non}.
In the early attempts, the CNN+LSTM structure~\cite{donahue2015long} achieved high performance by integrating two-dimensional convolutional layers, which has been popularly employed in image-related tasks, and a LSTM model targeting sequence data.
On the other hand, three-dimensional (3D) convolutional layers that utilize features in both spatial and temporal dimensions have been proposed \cite{ji20123d,tran2015learning}.
Another approach to deal with videos is to employ two CNNs simultaneously (known as two-stream networks), where they process original RGB frames (to exploit spatial features) and their optical flows (to exploit temporal features), respectively \cite{simonyan2014two}.
These two approaches are sometimes combined to further improve the performance of action recognition~\cite{carreira2017quo}.
In recent years, more advanced deep action recognition models have been developed.
One of the widely used approaches is an inflated three-dimensional (I3D) network~\cite{carreira2017quo,wang2018non}, which is a fine-tuned version of a pre-trained image classification model by inflating two-dimensional kernels of the convolutional layers to three-dimensional.
The two-stream approach also has been extended to a method named SlowFast~\cite{feichtenhofer2019slowfast}, which takes video data having different temporal resolutions (i.e., frame rates) on each of the stream networks as inputs.
Another trend in recent researches on action recognition is to employ kernel factorization (e.g., interaction-reduced channel-separated network (ir-CSN)~\cite{tran2019video}) for reducing computational complexity.

\subsection{Adversarial attack}
	
It has been shown that deep learning-based image classification models are highly vulnerable to adversarial attacks under a white-box scenario.
Szegedy \etal~\cite{szegedy2013intriguing} proposed an optimization-based attack method to minimize the amount of input perturbation that can change the classification result of a given model.
Goodfellow \etal~\cite{goodfellow2014explaining} developed the fast gradient sign method (FGSM), which calculates perturbation from the sign of the gradients obtained from a given model.
Kurakin \etal~\cite{kurakin2016adversarial} extended FGSM to an iterative approach, which is called iterative FGSM (I-FGSM) and showed higher fooling rates of the attack than FGSM.
While those methods find perturbation in the whole region of a given input image, Su \etal~\cite{su2019one} showed the feasibility of one-pixel attack, which tries to find perturbation of only one pixel to fool deep image classifiers.

The vulnerability of deep learning models has been further evaluated via several advanced methods beyond finding perturbation for each input image.
Liu \etal~\cite{liu2016delving} investigated transferability of perturbation, which is to examine whether a perturbation found for a model can also work for another model.
Moosavi-Dezfooli \etal~\cite{moosavi2017universal} found image-independent universal perturbation that can be applied to any images to fool a given classifier.

\subsection{Adversarial attack on action recognition}
There are a few studies on white-box adversarial attack of action recognition models.
Li \etal~\cite{li2018adversarial} developed an adversarial attack for the convolutional 3D (C3D) model~\cite{tran2015learning} by employing a generative adversarial network (GAN).
Wei \etal~\cite{wei2019sparse} proposed an optimization-based method to generate adversarial perturbation for LSTM-based models.
Pony \etal~\cite{naeh2020patternless} developed a method that changes the overall color of each frame in a given video clip to obtain a flickering perturbation.
However, these methods have been verified only on traditional action recognition models. Furthermore, they add perturbation to multiple frames of a given video, which may be visible to human observers.
\\[-0.6\baselineskip]

\section{Analysis of structural vulnerability}
\label{sec:3}

In this section, we show the existence of the structural vulnerability of action recognition models.
For this, a single frame of a video sequence is perturbed by I-FGSM~\cite{kurakin2016adversarial} and uniform random noise, and the recognition performance is examined for each frame.
Then, we analyze what factors cause such vulnerability.
\\[-0.6\baselineskip]

\begin{figure*}
	\centering
	\begin{subfigure}[b]{0.33\textwidth}
		\centering
		\includegraphics[width=\textwidth]{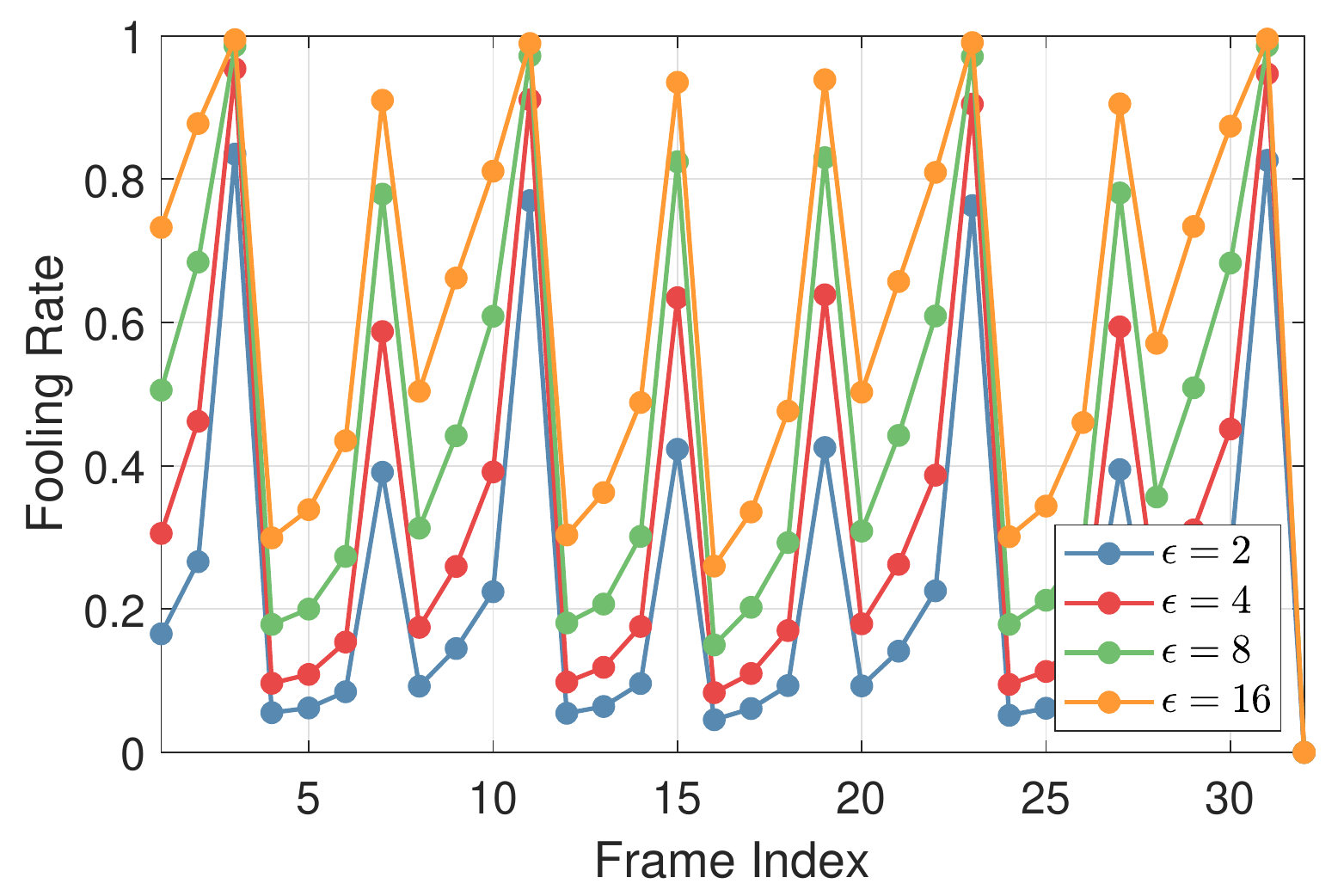}
		\caption{I3D}
	    \label{fig:property_I3D}
	\end{subfigure}
	\begin{subfigure}[b]{0.33\textwidth}
		\centering
		\includegraphics[width=\textwidth]{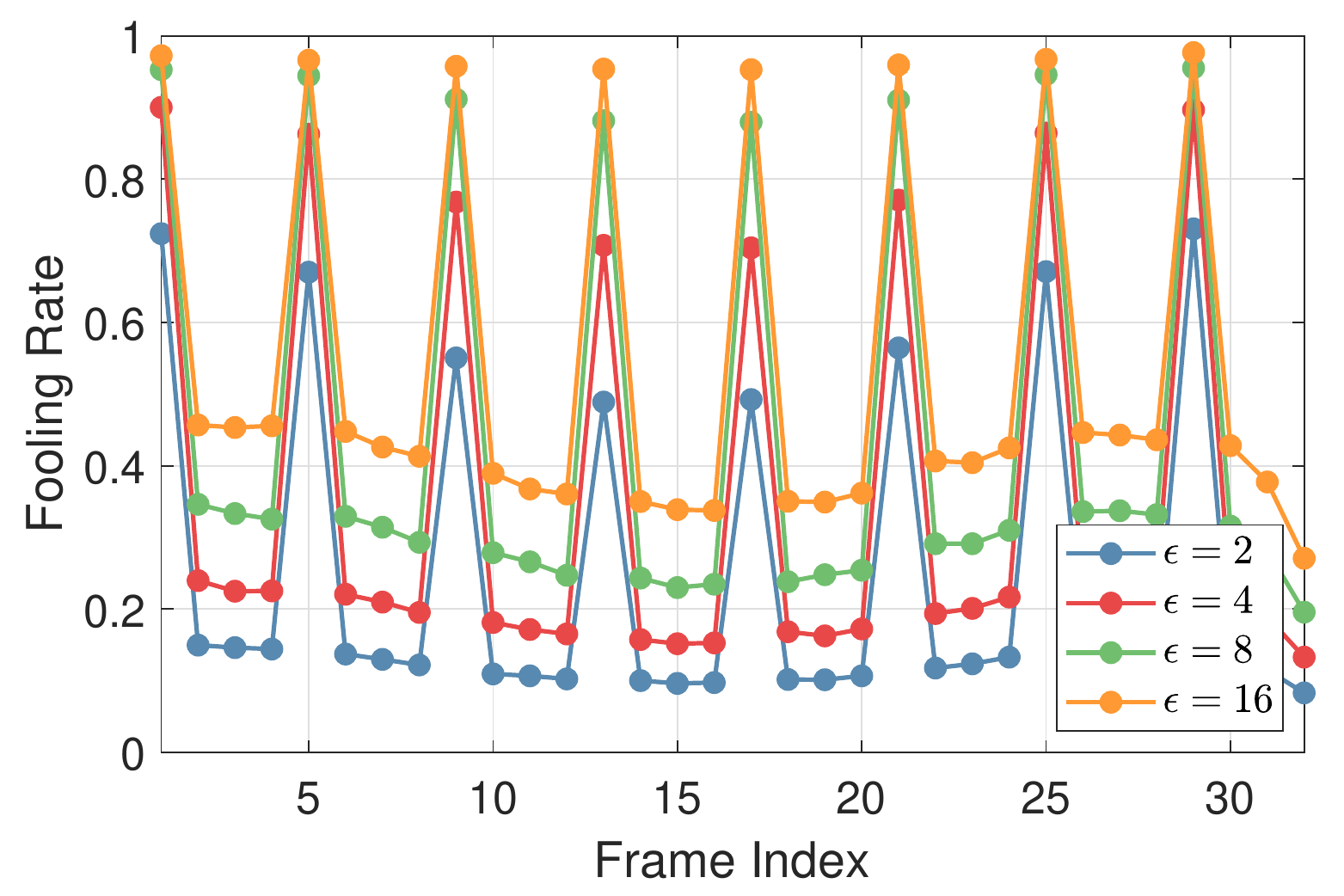}
		\caption{SlowFast}
	    \label{fig:property_SlowFast}
	\end{subfigure}
	\begin{subfigure}[b]{0.33\textwidth}
		\centering
	    \includegraphics[width=\textwidth]{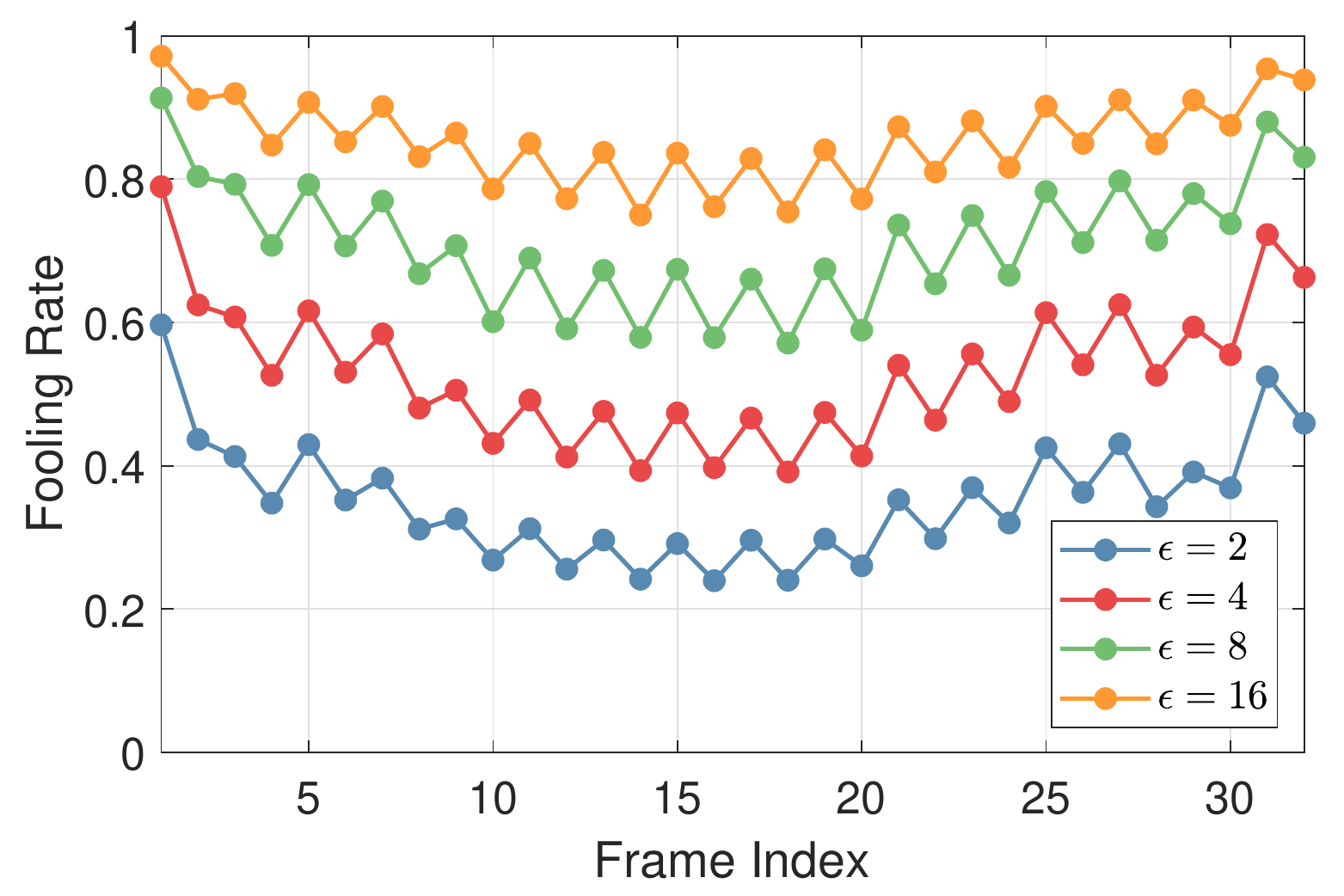}
	    \caption{ir-CSN}
	    \label{fig:property_ir-CSN}
	\end{subfigure}
	\vspace{-0.6cm}
	\caption{Fooling rates of perturbing a single frame by I-FGSM depending on the frame index.}
	\label{fig:property}
\end{figure*}

\begin{figure}
	\centering
	\includegraphics[width=0.85\columnwidth]{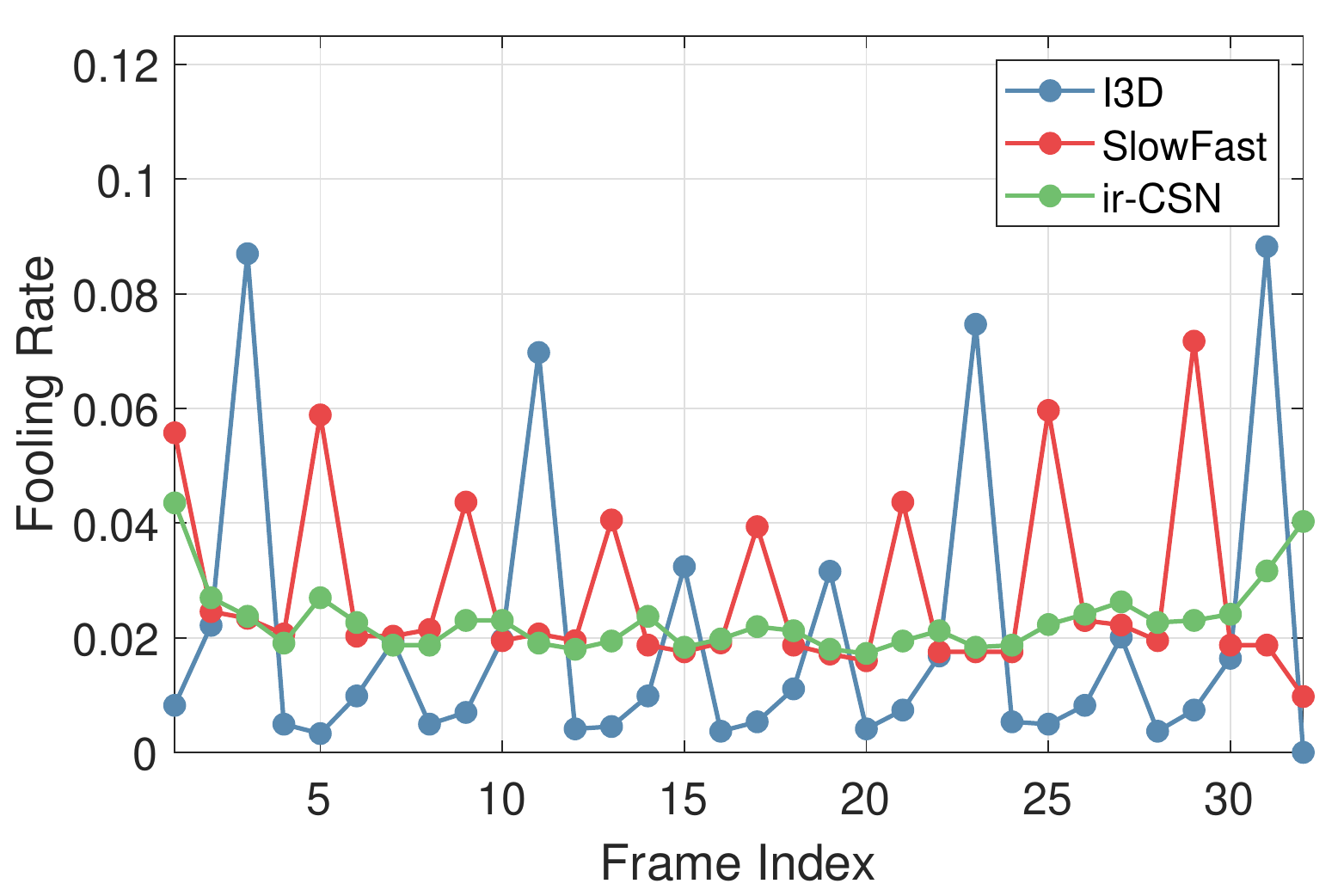}
	\vspace{-0.2cm}
	\caption{Fooling rates of perturbing a single frame by uniform random noise depending on the frame index.}
	\label{fig:random_noise}
\end{figure}

\subsection{Analysis methods}
\label{sec:3_1}

\paragraph{Using I-FGSM.}
We use I-FGSM~\cite{kurakin2016adversarial} to perturb a frame, which is one of the widely used strong adversarial attack methods.

This method iteratively finds a perturbation of the $i$-th frame in a video clip as follows.
Let $\mathrm{X}^{0} = \{ \mathrm{X}^{0}(1), \mathrm{X}^{0}(2), ..., \mathrm{X}^{0}(T) \}$ denote the original video clip (having $T$ frames) that is correctly classified as ${y}$ by a target action recognition model $M(\cdot)$, i.e., $M ( \mathrm{X}^{0} ) = {y}$.
The attack aims to find an attacked version of a video clip $\mathrm{X}$ from $\mathrm{X}^{0}$, where only the $i$-th frame $\mathrm{X}^{0}(i)$ is changed to an attacked one $\mathrm{X}(i)$ containing inconspicuous perturbation.
To find $\mathrm{X}(i)$, the I-FGSM update rule is applied, by finding the adversarial frame at iteration $n+1$, $\mathrm{X}^{n+1}(i)$, from the frame at the previous iteration, $\mathrm{X}^{n}(i)$, iteratively:
\begin{equation}
\begin{split}
	& \widetilde{\mathrm{X}}^{n+1}(i) = \\
	& \mathrm{Clip}_{0, 255} \Big( \mathrm{X}^{n}(i) + \frac{\epsilon}{N}~\mathrm{sgn}\big( \nabla_{\mathrm{X}^{n}(i)}{ J( \mathrm{X}^{n}, {y} ) } \big) \Big),
\end{split}
\end{equation}
\begin{equation}
	\mathrm{X}^{n+1}(i) = \mathrm{Clip}_{-\epsilon, \epsilon} ( \widetilde{\mathrm{X}}^{n+1}(i) - \mathrm{X}^{0}(i) ) + \mathrm{X}^{0}(i),
\end{equation}
where $\epsilon$ regulates the amount of perturbation to be added, $\mathrm{sgn}(\cdot)$ is the sign function, $\nabla_{\mathrm{X}^{n}(i)}{ J( \mathrm{X}^{n}, {y} ) }$ is the gradient of the target frame for the loss function $J( \mathrm{X}^{n}, {y} )$, and
\begin{equation}
	\mathrm{Clip}_{a, b}(\mathrm{X}) = \mathrm{min} \big( \mathrm{max} ( \mathrm{X}, a ), b \big).
\end{equation}
After $N$ iterations, the final adversarial video clip is obtained by $\mathrm{X} = \mathrm{X}^{N}=\{\mathrm{X}^0(1),\mathrm{X}^0(2),...,\mathrm{X}^N(i),\mathrm{X}^0(i+1),...,\mathrm{X}^0(T)\}$.
We expect that the model outputs a wrong prediction when $\mathrm{X}$ is inputted (i.e., $M ( \mathrm{X} ) \neq {y}$). 
\\[-1.8\baselineskip]

\paragraph{Using uniform random noise.} 
We use uniform random noise within [-64, 64] as perturbation, which is injected to a certain frame.
This type of perturbation is tested in order to understand the structural vulnerability.
Furthermore, since generating random noise perturbation is computationally efficient, it can be used to identify vulnerable frame indices for a given action recognition model.
\\[-0.6\baselineskip]

\subsection{Experimental setup}
\label{sec:3_2}

\paragraph{Dataset and models.}
We use Kinetics-400~\cite{kay2017kinetics}, which is one of the widely used large-scale benchmark datasets for action recognition.
From the test set of Kinetics-400, we randomly choose ten videos for each class.
Therefore, a total of 4000 videos are chosen to evaluate the fooling rates of the attack methods.
As target action recognition models, we consider three state-of-the-art models having various model structures, including I3D~\cite{wang2018non}, SlowFast~\cite{feichtenhofer2019slowfast}, and ir-CSN~\cite{tran2019video}.
These models have shown outstanding recognition performance on the Kinetics-400 dataset.
We employ the pre-trained models available on MMAction2~\cite{2020mmaction2}, which is the open-sourced repository that provides testing tools for the aforementioned action recognition models.
Among the variants of the SlowFast models, we use the $8 \times 8$ SlowFast in the implementation of MMAction2.
\\[-1.8\baselineskip]

\paragraph{Implementation details}
We conduct the I-FGSM method with various hyperparameters.
For attacking a single frame, we set the number of iterations to $N \in \{30, 50, 100\}$ and the amount of perturbation to $\epsilon \in \{2, 4, 8, 16\}$.
We only report the case of $N = 30$, which empirically found to be sufficient to attack the target models.
\\[-0.6\baselineskip]

\subsection{Vulnerability of action recognition models}
\label{sec:3_3}

Figures \ref{fig:property} and \ref{fig:random_noise} show the fooling rates of the two types of perturbation for the three models. 
Surprisingly, we observe the existence of vulnerable frame indices (or, shortly, vulnerable frames) showing significantly higher fooling rates than the others, especially in the cases of I3D and SlowFast.
We also confirm that those vulnerable frames are observed periodically.
Specifically, I3D and SlowFast have vulnerable frames at $i\in \{3, 7, 11, 15, 19, 23, 27, 31\}$ and $i\in \{1, 5, 9, 13, 17, 21, 25, 29\}$, respectively. 
The ir-CSN model does not show such a trend.
Compared to the other two models, it exhibits relatively high vulnerability overall, with relatively small variations between frames.
These observations hold consistently across different values of $\epsilon$.

Among the vulnerable frames, the most vulnerable frames for I3D, SlowFast, and ir-CSN are the 31st, 29th, and 1st frames, respectively.
It can be seen that even by adding uniform random noise, the most vulnerable frame can be identified.
Note that while Figure \ref{fig:random_noise} is obtained using all the video clips, we found that only 100 randomly chosen clips were enough to discover the most vulnerable frames.

\begin{figure}
	\centering
	\begin{subfigure}[b]{1.0\columnwidth}
		\centering
		\includegraphics[width=\columnwidth]{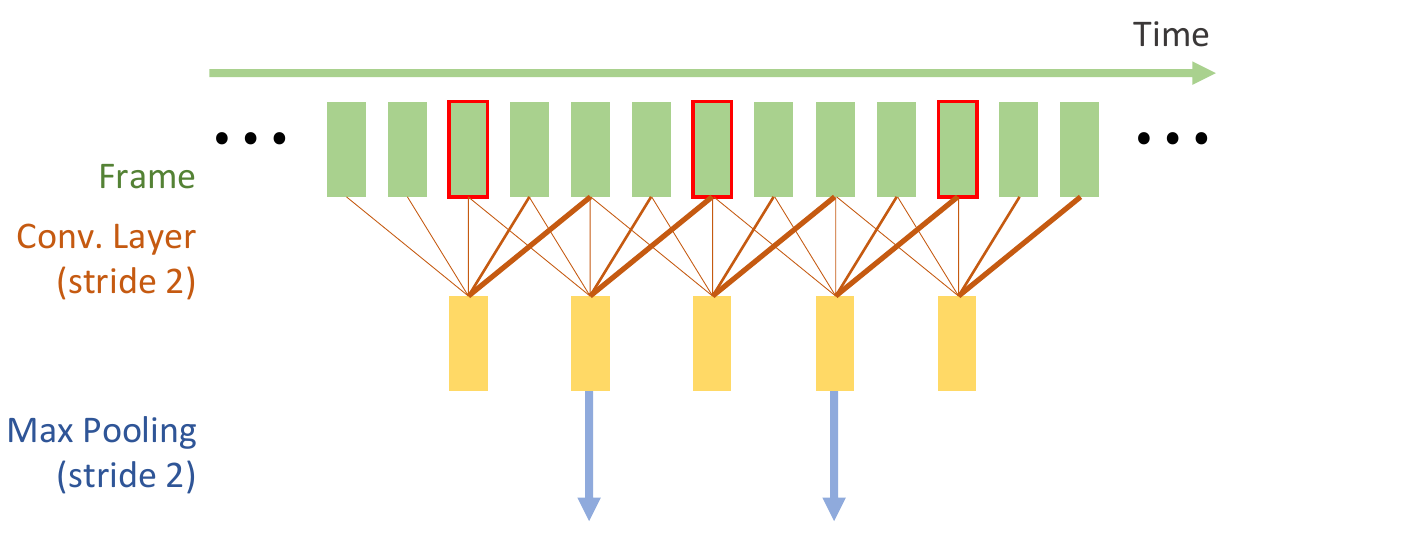}
		\caption{I3D}
		\label{fig:structure property1}
	\end{subfigure}
	\begin{subfigure}[b]{1.0\columnwidth}
		\centering
		\includegraphics[width=\columnwidth]{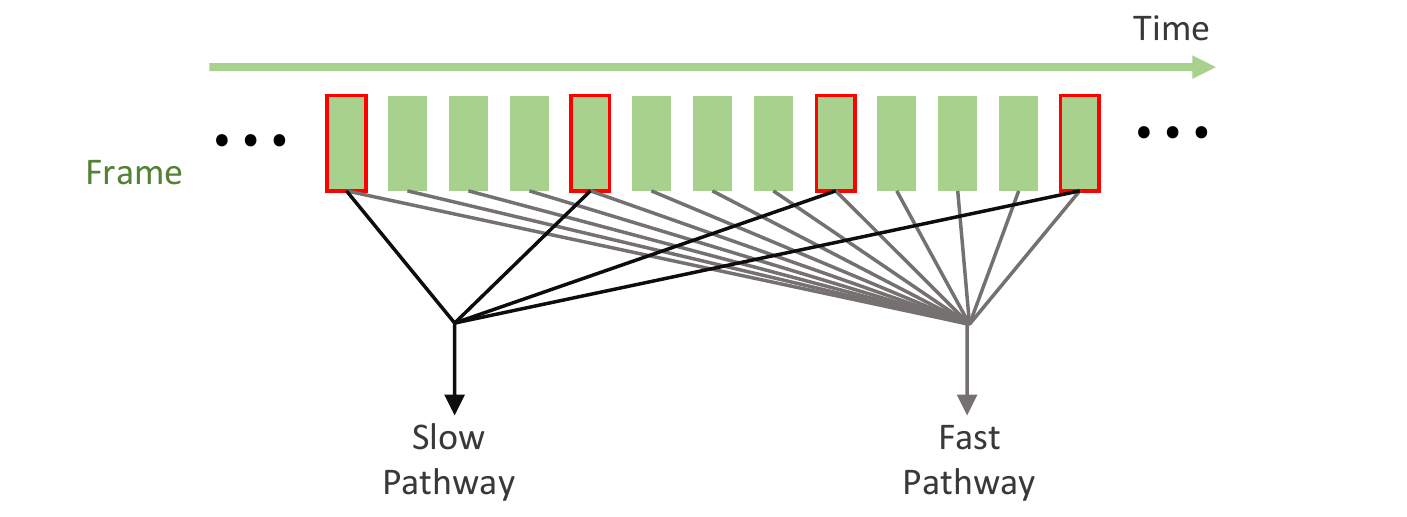}
		\caption{SlowFast}
		\label{fig:structure property2}
	\end{subfigure}
	\vspace{-0.5cm}
	\caption{Illustration of structural vulnerability. The frames marked with red boxes correspond to the vulnerable frames showing high fooling rates in Figure~\ref{fig:property}.}
	\label{fig:structure property}
\end{figure}

We discuss the causes of these interesting observations by analyzing the structural properties of the models, which are illustrated in Figure \ref{fig:structure property}.
\\[-1.8\baselineskip]

\begin{figure*}[t]
	\centering
	\begin{subfigure}[b]{0.33\textwidth}
		\centering
		\includegraphics[width=\textwidth]{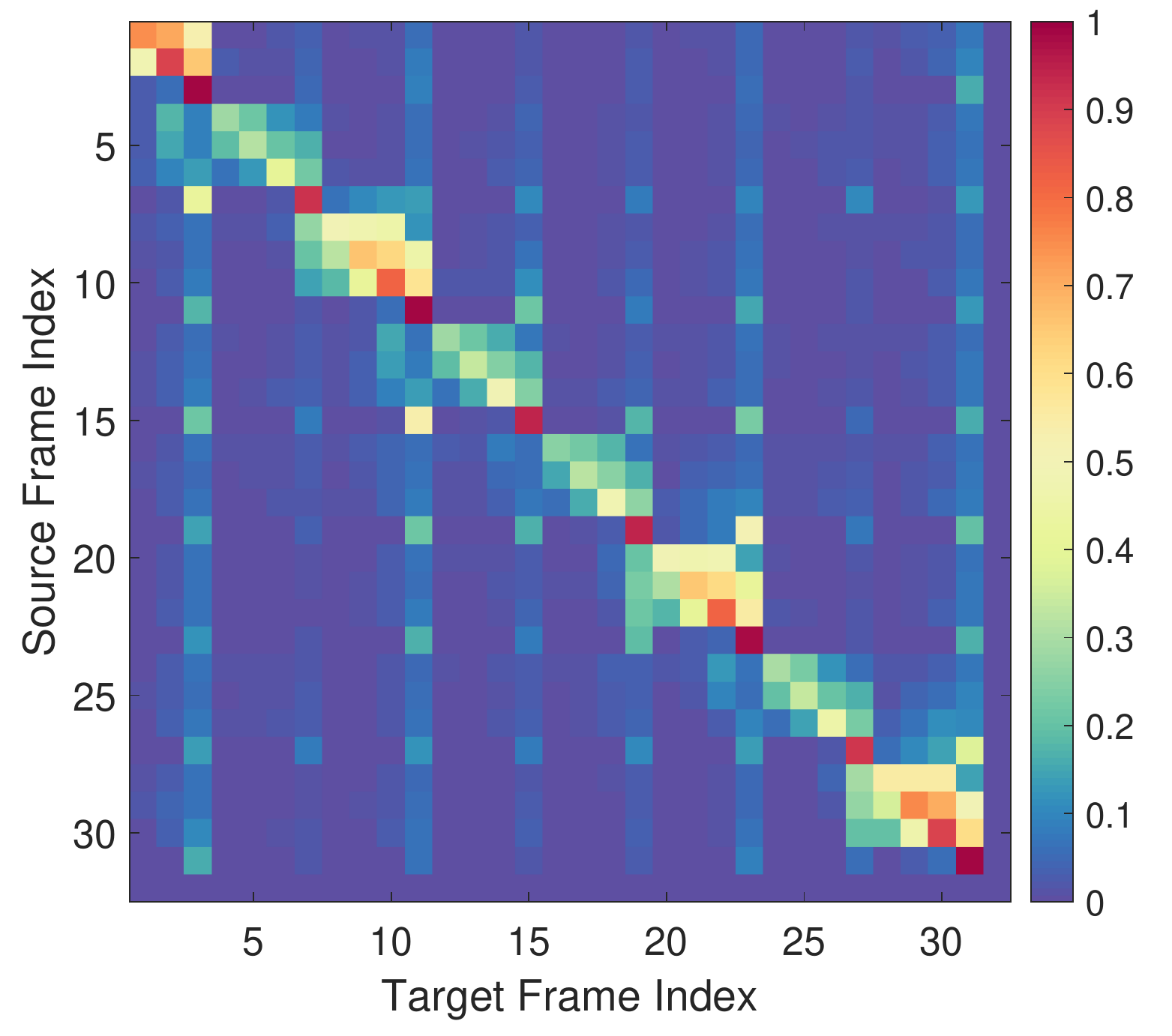}
		\caption{I3D}
	\end{subfigure}
	\begin{subfigure}[b]{0.33\textwidth}
		\centering
		\includegraphics[width=\textwidth]{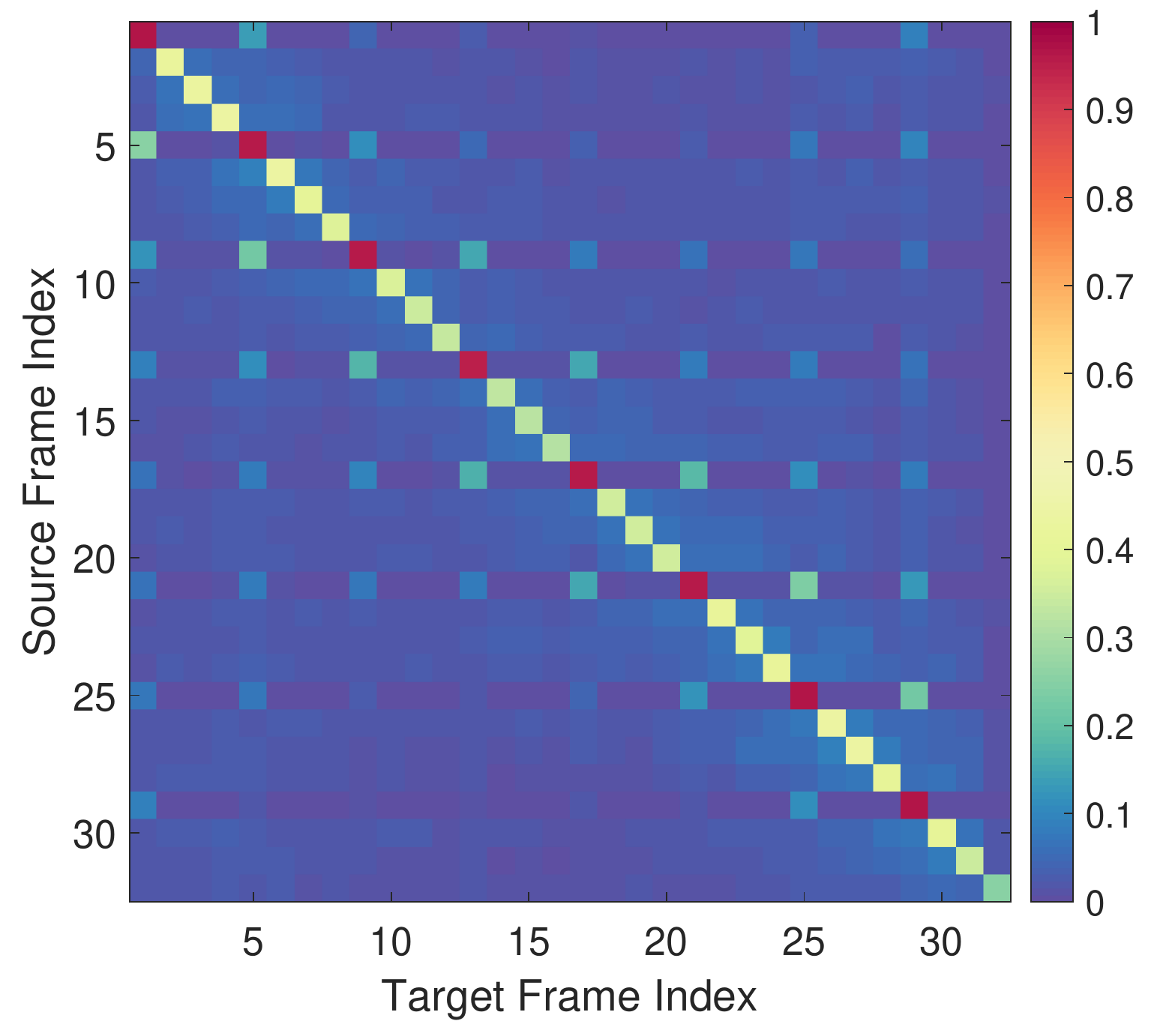}
		\caption{SlowFast}
	\end{subfigure}
	\begin{subfigure}[b]{0.33\textwidth}
		\centering
		\includegraphics[width=\textwidth]{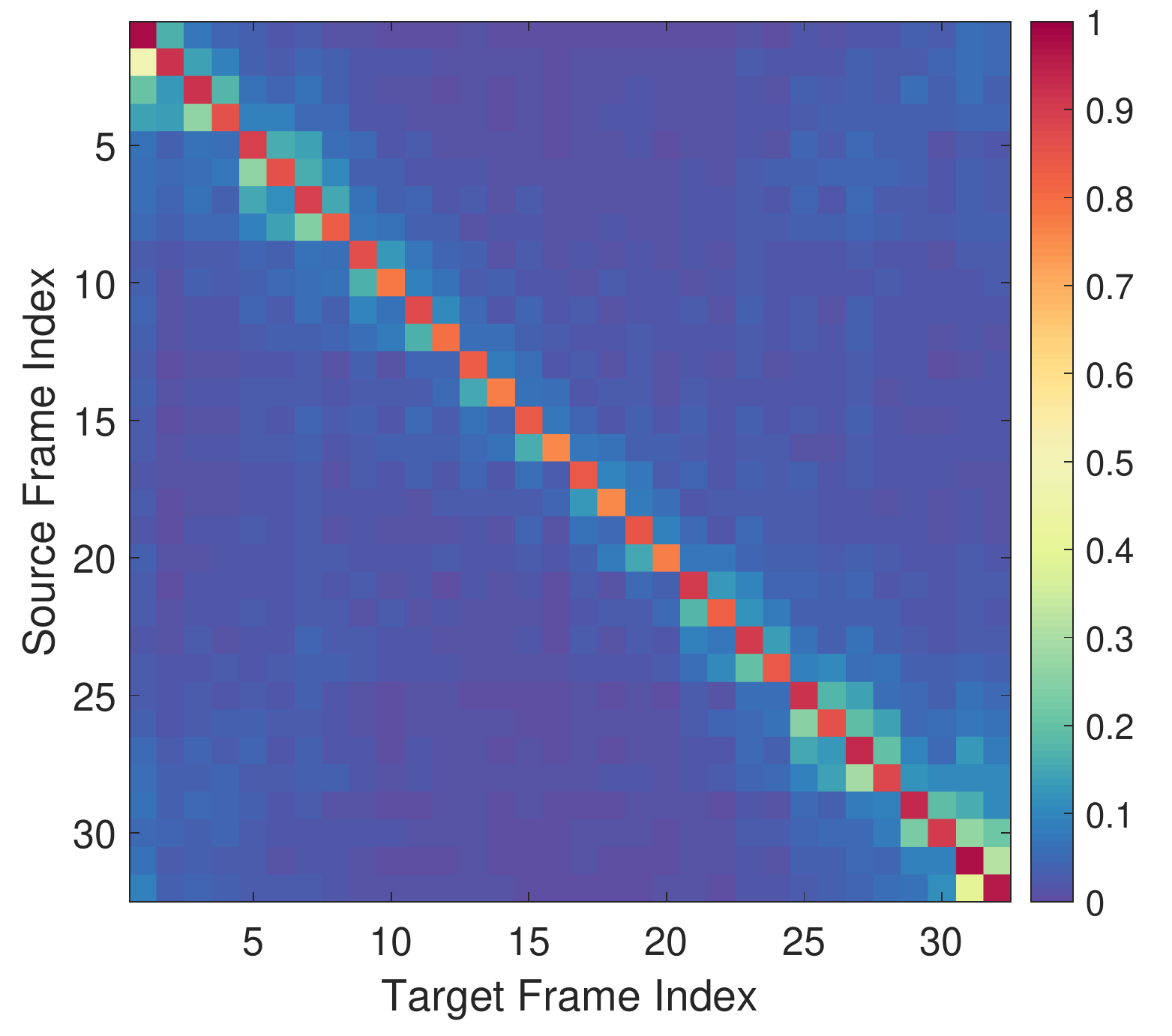}
		\caption{ir-CSN}
	\end{subfigure}
	\vspace{-0.6cm}
	\caption{Transferability between different frames in each model. The value at $(x, y)$ means the fooling rate for the $x$-th frame using the perturbation generated for the $y$-th frame. Red and blue colors indicate high and low values, respectively.}
	\label{fig:transferability}
\end{figure*}

\paragraph{I3D.}
As illustrated in \figurename~\ref{fig:structure property1}, the I3D model \cite{wang2018non} initially extracts features from a given video clip via a convolutional layer having a kernel size of $5 \times 7 \times 7$ and a temporal stride of 2.
Then, they are processed by a max-pooling layer having a kernel size of $1 \times 3 \times 3$ and a temporal stride of 2.
Through this process, the video clip having 32 frames is contracted to eight frames of features, i.e., an effective temporal stride becomes 4. 
And, there may exist asymmetric information extraction through the two layers, i.e., when the weights of the convolutional layer have different values across the temporal dimension, the layer's output relies more on the information at specific frames and less on the information at the other frames among the five frames entering to the kernel, which is more emphasized through the max-pooling layer. 
In the case of the pre-trained I3D model used in our experiment, the average magnitude of the weights of the first convolutional layer is measured as 0.01, 0.01, 0.02, 0.03, and 0.10, which means that the layer mostly relies on the information in the fifth frame among the five input frames (indicated by thicker lines in~\figurename~\ref{fig:structure property1}). 
By these two mechanisms (an effective stride of 4 and the asymmetric information extraction), the perturbation inserted at $i \in \{3, 7, 11, 15, 19, 23, 27, 31\}$ easily attacks the model (the frames marked with red boxes in~\figurename~\ref{fig:structure property1}), while the perturbation at the other frames does not.
\\[-1.8\baselineskip]

\paragraph{SlowFast.}
SlowFast~\cite{feichtenhofer2019slowfast} is a two-stream model, which includes a fast pathway and a slow pathway.
Since the fast pathway uses all of the 32 frames and the slow pathway takes only every fourth frame, only eight frames are simultaneously used by both pathways, which are marked with red boxes in Figure~\ref{fig:structure property2}.
We find that the highly vulnerable frames ($i\in \{1, 5, 9, 13, 17, 21, 25, 29\}$) exactly match the frames used by both pathways.
The other frames are processed only through the fast pathway and thus the perturbation on those frames is not so successful. 
Note that the fast pathway has the structure of I3D but the temporal stride is 1 unlike the original I3D, so the above-observed periodic pattern does not appear here. 
\\[-1.8\baselineskip]

\paragraph{ir-CSN.}
The ir-CSN~\cite{tran2019video} model used in our study is based on ResNet-152, which is deeper than the other two models using ResNet-50.
As aforementioned, the ir-CSN model is relatively vulnerable across all frames; for example, when $\epsilon$ is 16, even the lowest fooling rate is 75.4\%.
The strides in the first convolutional layer and the first pooling layer are 1, thus all input frames are treated evenly in these layers. 
Therefore, the vulnerability is rather similar across all frames unlike I3D and SlowFast.
It is also observed that the fooling rate increases around the two edge frames, which seems to be because zero paddings highlight the perturbation in the edge frames.
\\[-1.8\baselineskip]

\paragraph{Transferability of perturbation.}
We investigate the transferability of perturbations between frames using I-FGSM, i.e., whether the perturbation generated for a frame can be also used for another frame location directly to attack the model.
Figure \ref{fig:transferability} shows that there exist pairs of source and target frame locations showing particularly high transferability, and the transferability patterns differ depending on the action recognition model.

For I3D and SlowFast, high transferability is achieved between relatively more vulnerable frames (shown in Figure~\ref{fig:property}).
The fooling rate by transferred perturbation is still higher than that by the uniform random noise attack.
This suggests that perturbations for vulnerable frames have common features making the model operate wrongly.

Furthermore, in the cases of I3D and ir-CSN, there is a relatively high level of transferability between adjacent frames. 
However, this is not the case in SlowFast, because the slow pathway does not take all adjacent frames but only every fourth frame.

\newcolumntype{b}{>{\hsize=1.8\hsize\centering\arraybackslash}X}
\newcolumntype{s}{>{\hsize=0.8\hsize \centering\arraybackslash}X}
\newcolumntype{n}{>{\hsize=1\hsize \centering\arraybackslash}X}
\renewcommand{\arraystretch}{1.2}

\begin{table}
	\centering
	\small
	\begin{tabularx}{\columnwidth}{b|ssssn}
		\toprule
		&\multicolumn{4}{c}{One frame attack}&\multirow{2}{*}{{Wei~\cite{wei2019sparse}}}\\
	    &$\epsilon$=2&$\epsilon$=4&$\epsilon$=8&$\epsilon$=16\\
		\midrule
		I3D~\cite{wang2018non}&0.83&0.95&0.99&1.00&0.81\\ 
		SlowFast~\cite{feichtenhofer2019slowfast}&0.73&0.90&0.96&0.98&0.72\\
		ir-CSN~\cite{tran2019video}&0.60&0.79&0.91&0.97&0.68  \\ \bottomrule
	\end{tabularx}
	\caption{Fooling rates of the one frame attack and the compared attacked method~\cite{wei2019sparse} on the three action recognition models. The one frame attack is conducted with various values of $\epsilon$.}
	\label{table:performs}
	\centering
\end{table}

\section{Vulnerability against one frame attack}
\label{sec:4}

In Section \ref{sec:3}, we could find the most vulnerable frames related to the structural vulnerability of an action recognition model.
In this section, we perform the \emph{one frame attack} under a white-box scenario, which applies the I-FGSM algorithm to only the most vulnerable frames, i.e., the 31st, 29th, and first frames for I3D, SlowFast, and ir-CSN, respectively.
We evaluate the performance of this attack with respect to two criteria: 1) the fooling rate and 2) the degree of inconspicuousness of perturbation.
It is demonstrated that by exploiting the structural vulnerability, the one frame attack can fool the models with high fooling rates and high invisibility in comparison to the existing attack methods.
\\[-0.6\baselineskip]

\subsection{Fooling rate}
\label{subsec:4_1}
Table \ref{table:performs} summarizes the fooling rates of the one frame attack for each case.
The attack achieves high fooling rates for all target models.
Especially, when $\epsilon$ is equal to or larger than 8, the fooling rates exceed 90\%.
The fooling rates are over 60\% even when $\epsilon$ is as small as 2. 

For comparison, we implement the attack method in \cite{wei2019sparse} to attack the deep action recognition models.
The fooling rate of this method is lower than that of the one frame attack except the case of ir-CSN with $\epsilon=2$.
Considering that the one frame attack with up to $\epsilon=16$ is fairly inconspicuous (see Section \ref{subsec:4_2}), the method in \cite{wei2019sparse} does not effectively capture the vulnerability of the action recognition models.
This is because it is designed to exploit the particular mechanism (temporal information propagation) of the LSTM-based models.
\\[-0.6\baselineskip]

\begin{figure*}
	\centering
	\includegraphics[width=1.0\textwidth]{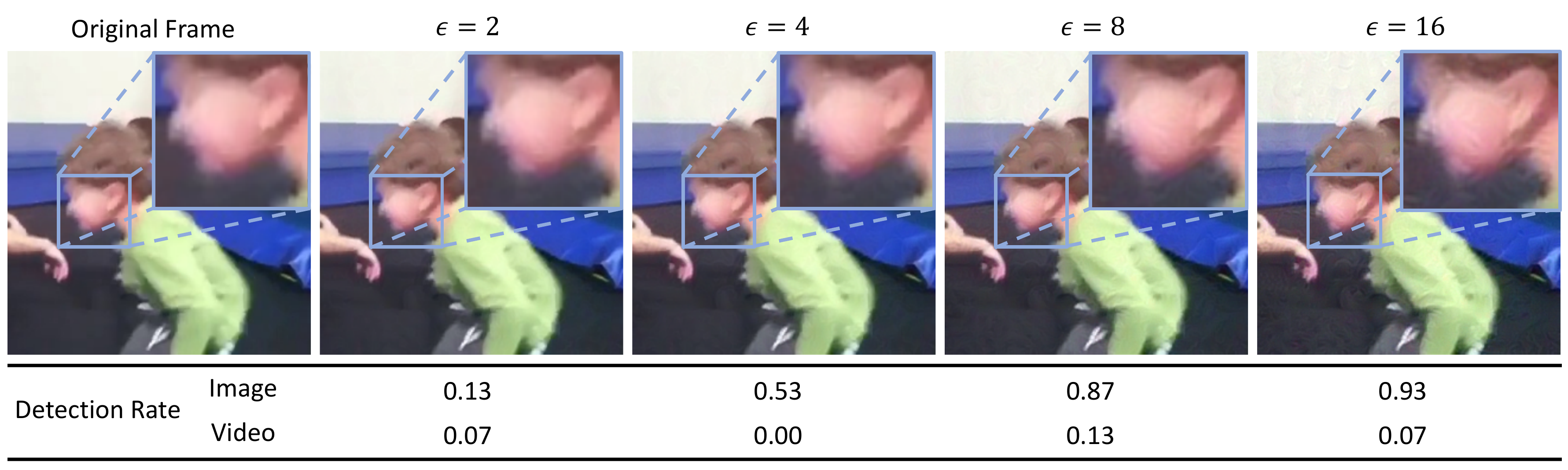}
	\vspace{-0.5cm}
	\caption{Visibility and the detection rate of the perturbations with $\epsilon = 2, 4, 8, 16$ for I3D.}
	\label{fig:adversarial ex}
\end{figure*}

\subsection{Inconspicuousness}
\label{subsec:4_2}
We conduct a subjective test to examine the level of inconspicuousness of the one frame attack.
We use 4 (the number of videos) $\times$ 4 ($\epsilon \in \{2, 4, 8, 16\}$) $\times$ 3 (the number of target models) $=48$ perturbed videos.
Fifteen participants are employed, which meets the required number of participants for subjective tests according to the recommendation ITU-R BT.500-13~\cite{union2012recommendation}.
Our test is based on the double-stimulus impairment scale (DSIS) method in~\cite{union2012recommendation}.
In other words,	the participants watch the original video and its perturbed version sequentially for three seconds each with a mid-gray image displayed between the two by following~\cite{union2012recommendation}.
The order of the video pair is randomly switched.
Then, the participants answer whether they notice the difference between the video pair.
We also include pairs of original videos to obtain a ``baseline'' detection performance of the participants.
The same procedure is repeated for the pairs of a perturbed frame and its original version where the exposure time of one image is set to two seconds.
For comparison, we implement the flickering attack \cite{naeh2020patternless} aiming at a high level of inconspicuousness where flickering perturbation is added to all frames to attack a video clip, and the resulting videos are also evaluated.

\begin{figure}
	\centering
	\includegraphics[width=0.85\columnwidth]{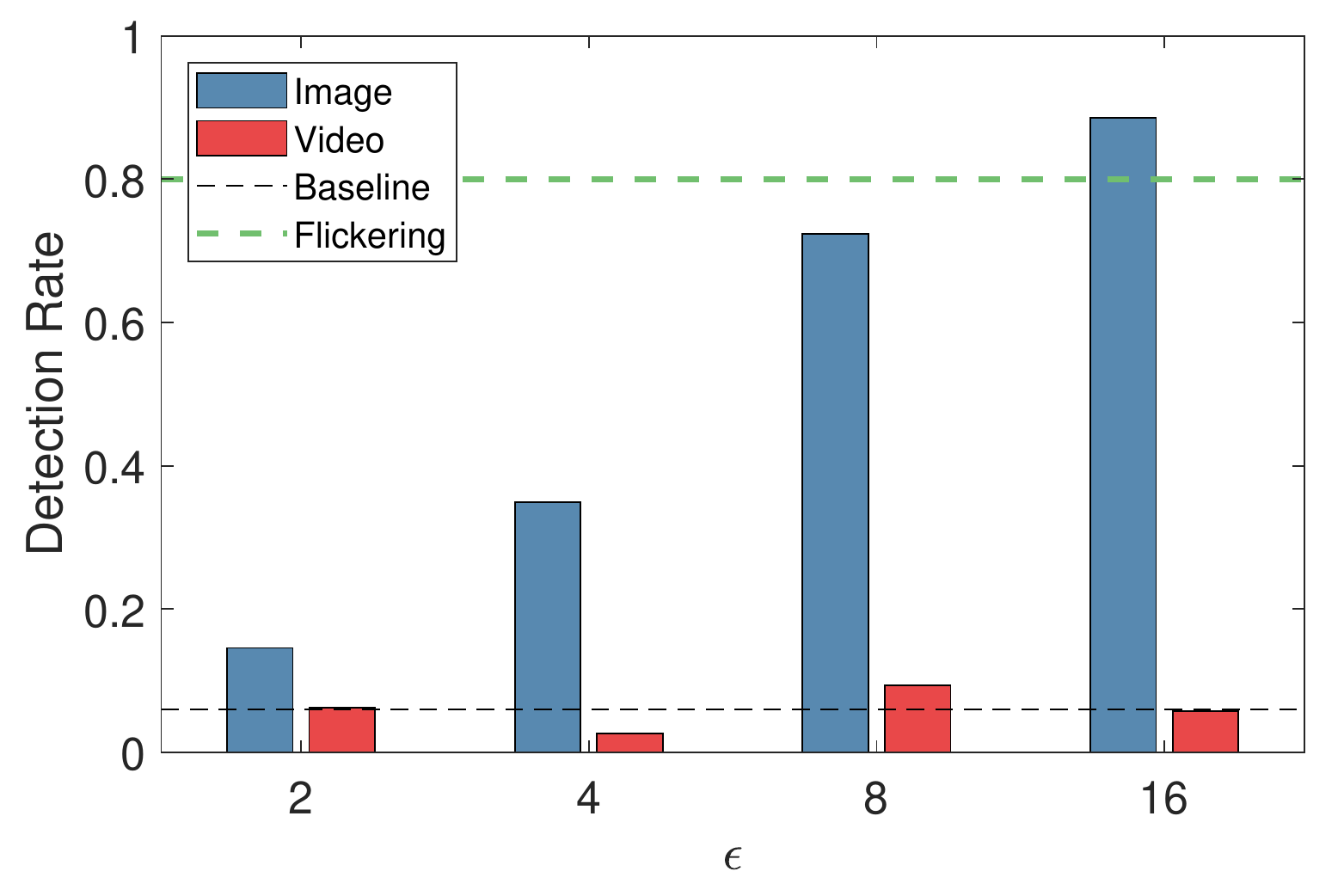}
	\vspace{-0.2cm}
	\caption{Results of the subjective test in terms of the detection rate. `Image' and `Video' correspond to the cases where the image frames perturbed by the one frame attack and the video clips containing the attacked frames are shown to the participants, respectively. `Baseline' indicates the erroneous detection rate for pairs of original video clips. `Flickering' means the detection rate for the perturbed videos by the flickering attack~\cite{naeh2020patternless}.}
	\label{fig:subjective test}
\end{figure}

\begin{figure*}
	\centering
	\begin{subfigure}[b]{0.33\linewidth}
	\includegraphics[width=\textwidth]{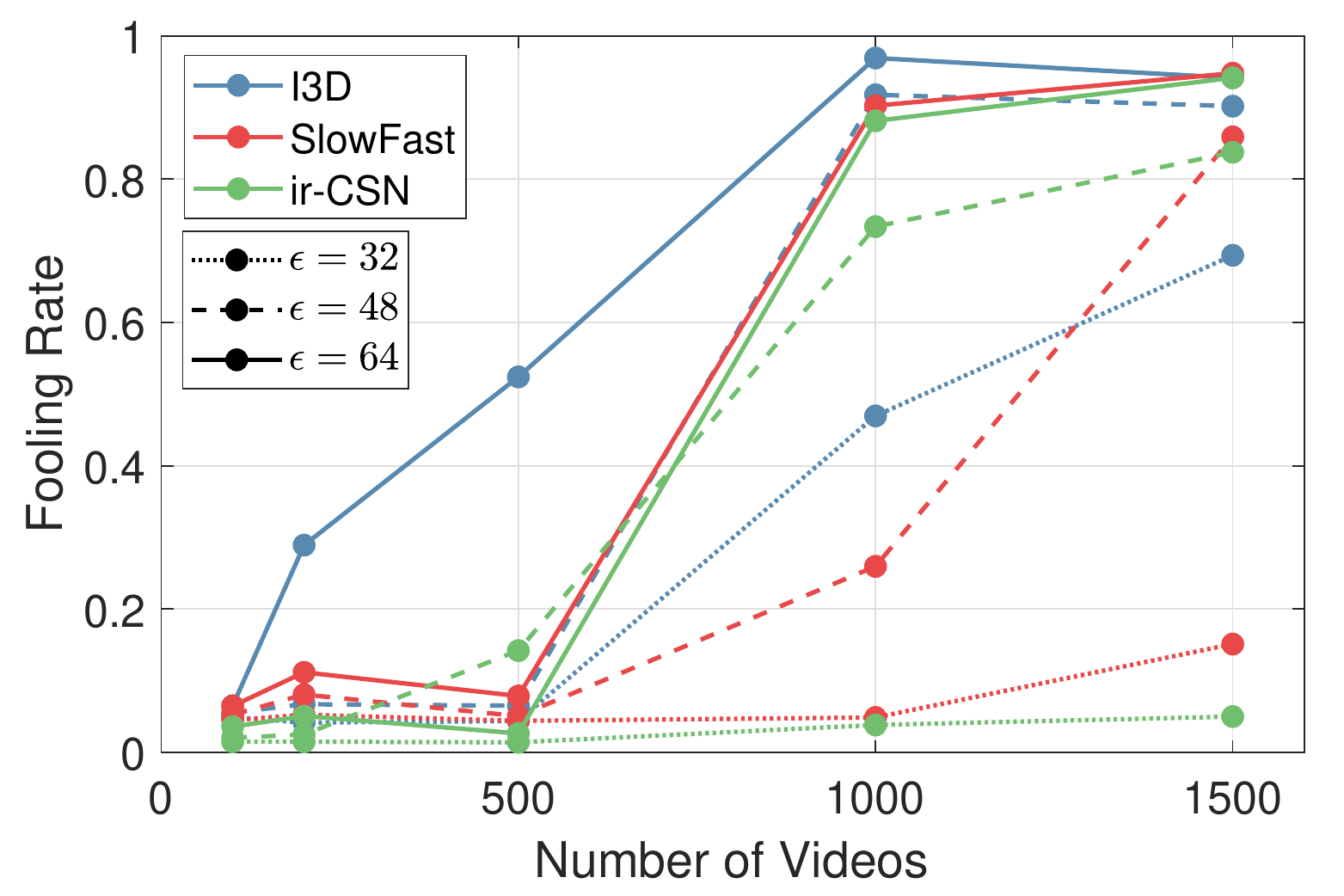}
	\caption{}
	\label{fig:univ_one_a}
	\end{subfigure}
	\begin{subfigure}[b]{0.33\linewidth}
	\includegraphics[width=\textwidth]{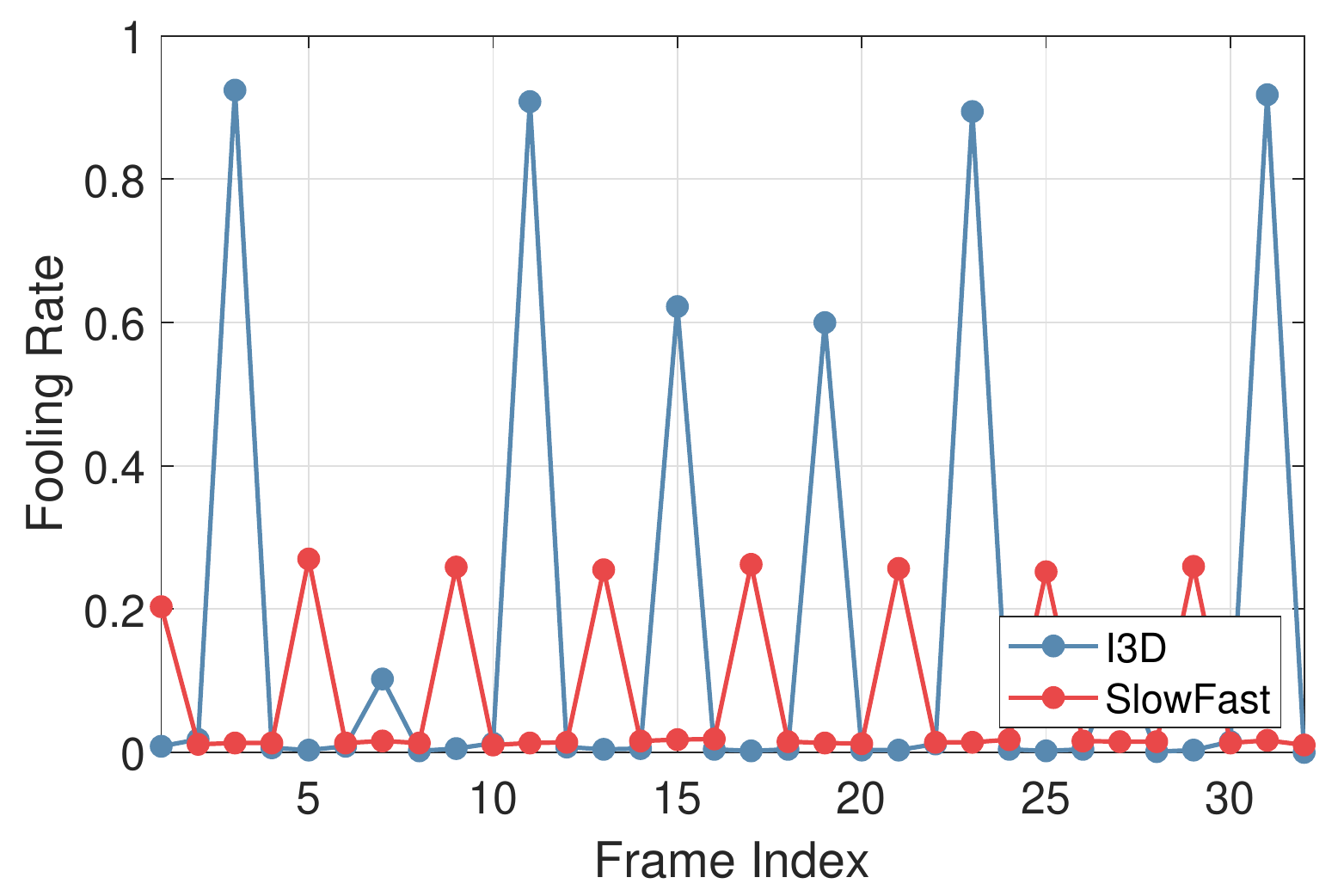}
	\caption{}
	\label{fig:univ_one_b}
	\end{subfigure}
	\begin{subfigure}[b]{0.33\linewidth}
	\includegraphics[width=\textwidth]{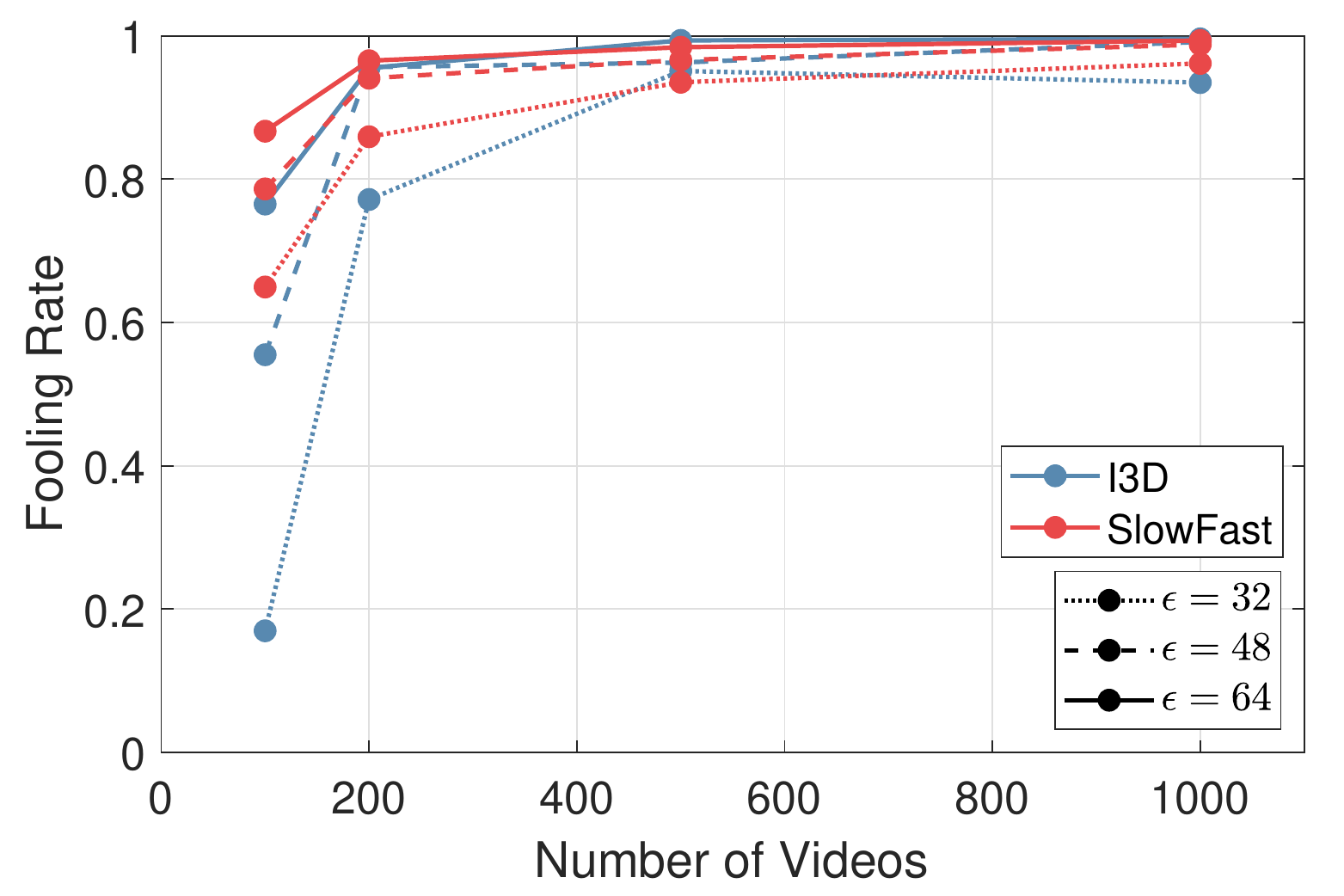}
	\caption{}
	\label{fig:univ_one_c}
	\end{subfigure}
	\caption{Results of the universal one frame attack. (a) Fooling rates with respect to the number of videos when the most vulnerable frame of each video clip is attacked. (b) Transferability of the universal perturbation found using the most vulnerable frame of each of 1500 videos to the other frames, where $\epsilon = 48$. (c) Fooling rates with respect to the number of videos when multiple vulnerable frames in each video clip are used to obtain the perturbation.}
	\label{fig:univ_one}
\end{figure*}

Figure \ref{fig:adversarial ex} shows examples of perturbed frames for I3D when we set $\epsilon \in \{2, 4, 8, 16\}$.
We also present the detection rates of these frames (when viewed as images) and the videos containing these frames (when viewed as vidoes).
The perturbation in the frames for high values of $\epsilon$ is easily found by the participants. 
As a result, we obtain higher detection rates when they are viewed as images than when they are viewed as videos.

Figure \ref{fig:subjective test} shows the overall results of the subjective test.
The frames perturbed by the one frame attack, when viewed as images, are relatively easily detected especially for large values of $\epsilon$. 
The detection rate of the images increases as $\epsilon$ increases, which is natural. 
However, the videos containing the perturbed frames are hardly detected, showing detection rates even lower than the erroneous detection rate (`baseline') for the pairs of the same original videos. 
In contrast, the flickering attack is easily detectable.
With these results, we can confirm the inconspicuousness of the one frame attack.
\\[-0.6\baselineskip]

\section{Vulnerability against universal one frame attack}
\label{sec:5}

In Section~\ref{sec:4}, the one frame attack finds the adversarial perturbation for each video clip and deteriorates the performance of the recognition models, showing the risk of the vulnerability.
In this section, we examine the possibility of universal attack~\cite{moosavi2017universal}, which is to find an video-agnostic perturbation that can affect any video clip for a target action recognition model.
In addition, we investigate the possibility of extending the one frame attack to a time-invariant universal attack, which assumes a real-time action recognition situation.
\\[-0.6\baselineskip]

\subsection{Attack method}
\label{sec:5_1}

The video-agnostic universal perturbation is obtained in a similar way to the method described by (1) and (2).
However, instead of the gradient for each video clip, the average gradient for $K$ videos is used in the sign function, i.e.,

\begin{equation}
    \label{eq:universal_attack_average_gradient}
	\mathrm{G}^{{n+1}} = {1\over{K}}\sum_{k=1}^K \sum_{i\in I} {\nabla_{\mathrm{X}_{k}(i)}}J(\mathrm{X}_k^n,y_k)
\end{equation}

\noindent where $I$ is the set of target frame indices that are used to find a universal perturbation, $\mathrm{X}_{k}(i)$ is the $i$-th frame of the $k$-th video clip, and $y_k$ is the ground-truth label of the $k$-th video clip.
\\[-0.6\baselineskip]

\subsection{Implementation details}
\label{sec: 5_2}

For finding universal perturbations, we consider two different frame sets $I$: the set of multiple highly vulnerable frames and the set of only the most vulnerable frame. 
We set $N = 100$ and $\epsilon \in \{32, 48, 64\}$.
Note that a larger number of iterations and larger values of $\epsilon$ are required to find universal perturbations compared to the case of video-specific perturbations, as also mentioned in the previous studies \cite{naeh2020patternless,wei2019sparse}.
Besides, finding a universal perturbation requires a high computational complexity, since the gradients need to be calculated from all target videos at every iteration.
Employing a larger number of videos to find a universal perturbation takes more time, but a higher fooling rate can be expected since the perturbation is found from more diverse videos.
To examine this, we vary the number of videos $(K)$ to find the universal perturbation, where $K \in \{100, 200, 500, 1000, 1500\}$.

To test the universal perturbation, we randomly choose additional 1000 videos from the Kinetics-400 dataset,
which are different from the videos used for generating the universal perturbation.
\\[-0.6\baselineskip]

\subsection{Results of universal attack}
\label{sec: 5_3}

\paragraph{Attack on the most vulnerable frame.}
We first find a universal perturbation from the most vulnerable frames found in Section~\ref{sec:3_3} (i.e., $I$ in (\ref{eq:universal_attack_average_gradient}) has only one frame index) for all $K$ video clips.
\figurename~\ref{fig:univ_one_a} shows fooling rates of this universal one frame attack with respect to the value of $K$.
The results show that increasing the number of videos is usually beneficial to achieve a higher fooling rate.
When 1500 videos are used, the attack with $\epsilon = 32$ achieves a fooling rate above 80\% for every target action recognition model, indicating the universal one frame attack is feasible.
\\[-1.8\baselineskip]

\paragraph{Transferability of universal perturbation.}
We also investigate the transferability of the universal perturbation found from the most vulnerable frames.
To do this, for every target frame index, we add the same universal perturbation to that frame and measure the fooling rate.
\figurename~\ref{fig:univ_one_b} shows the results for I3D and SlowFast.
It is shown that the vulnerable frame indices found in Section~\ref{sec:3_3} are also highly vulnerable against the transferred universal perturbation, i.e., the universal perturbation is highly transferrable among the vulnerable frames.
\\[-1.8\baselineskip]

\paragraph{Attack on multiple vulnerable frames.}
Although the one frame attack can find a strong universal perturbation even when only the most vulnerable frame is exploited, more powerful universal perturbation can be found using multiple vulnerable frames.
To evaluate this, we find a universal perturbation using all the vulnerable frames found in Section~\ref{sec:3} (i.e., $I$ in (\ref{eq:universal_attack_average_gradient}) has multiple frame indices).
\figurename~\ref{fig:univ_one_c} shows the results when the found perturbation is injected to the most vulnerable frames.
When compared to the result of the universal attack found using only the most vulnerable frame (\figurename~\ref{fig:univ_one_a}), this result shows that the universal perturbation found from multiple vulnerable frames is more powerful.
For example, when 500 or 1000 videos are employed to find the universal perturbation, the fooling rates for both I3D and SlowFast exceed 90\% in any values of $\epsilon$ in \figurename~\ref{fig:univ_one_c}.
In addition, sufficiently strong universal perturbations can be found even when a small number of videos are employed.
For instance, the fooling rates for all cases when $\epsilon \in \{48, 64\}$ exceed 90\% when only 200 videos are employed.

\begin{figure}
	\centering
	\includegraphics[width=0.85\linewidth]{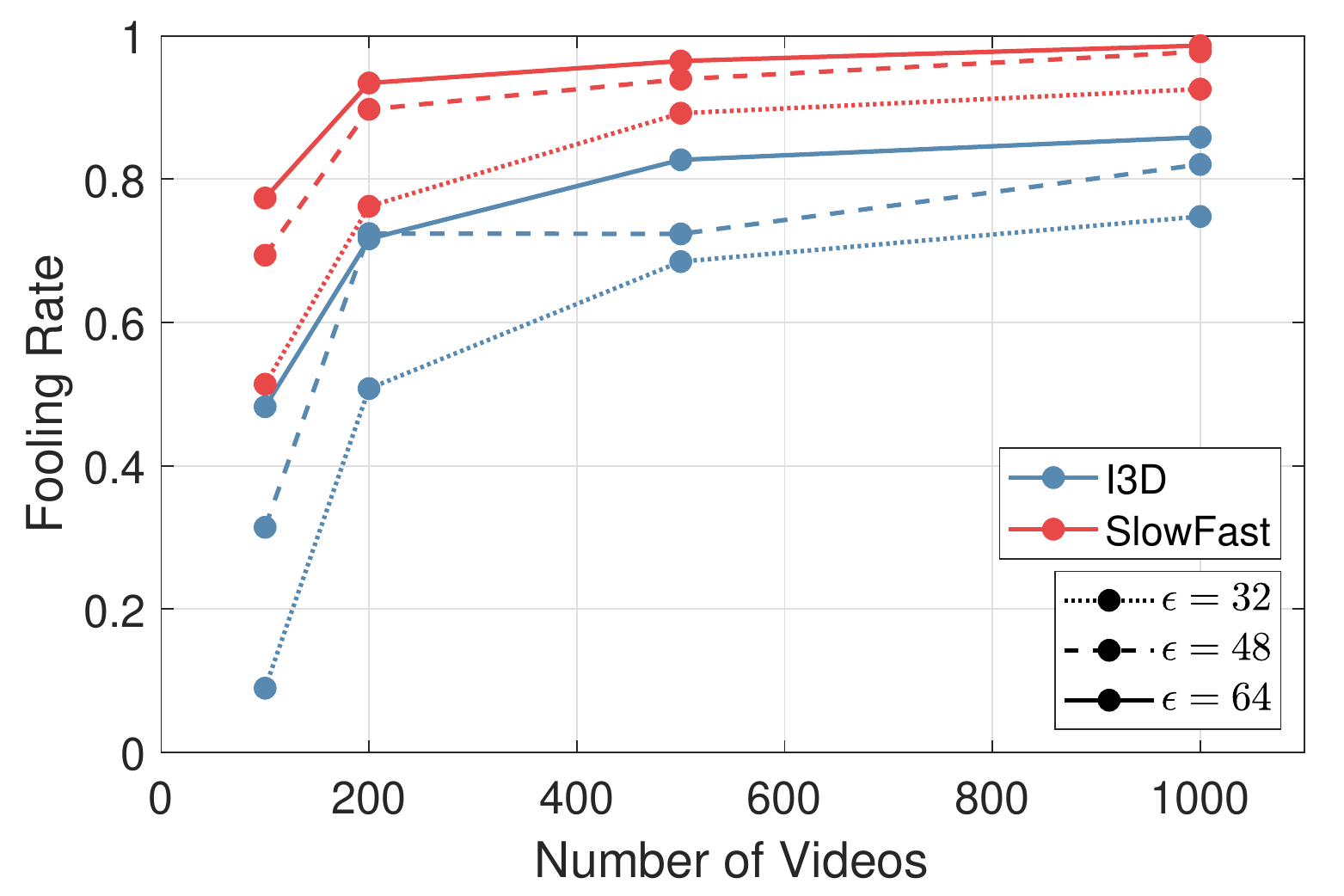}
	\caption{{Fooling rates of the time-invariant universal attack with respect to the number of videos.}}
	\label{fig:univ_v}
\end{figure}

\section{Vulnerability against time-invariant universal attack}
\label{subsec:6}

We additionally consider a real-time recognition scenario, where video data is continuously generated and a video clip is repeatedly chosen for recognition from the generated video sequence using a sliding window \cite{li2018adversarial}.
Two challenges arise in this scenario.
First, there may not be enough time to generate perturbations specific to the inputted video.
Therefore, a universal perturbation computed a priori needs to be used to ensure real-time operation.
Second, there may exist an unknown temporal offset between the video clip chosen for recognition and the video clip that the attacker observes.
An attack method that can deal with this scenario is referred to as a time-invariant attack \cite{naeh2020patternless}.
In this section, we examine the feasibility of the universal one frame attack as a time-invariant attack.

As shown in Section \ref{sec:3_3}, vulnerable frames appear periodically for I3D and SlowFast.
By taking advantage of this, a time-invariant universal attack can be designed as follows.
Let $P$ denote the period of vulnerable frames and $L$ the length of a video clip, which are 4 and 32 for both I3D and SlowFast, respectively.
Then, we add a pre-computed universal one frame perturbation, which is generated by attacking on multiple vulnerable frames in Section \ref{sec: 5_3}, to $P$ frames in every $L$ frames in such a way that the frame index of each of the $P$ frames corresponds to a distinct remainder when divided by $P$.
For instance, the 1st, 10th, 19th, and 24th frames, whose remainders are 1, 2, 3, and 0 when divided by 4, respectively, are perturbed, which is repeated for the next 32 frames.
This ensures that one of the $P$ perturbed frames always corresponds to a vulnerable frame index no matter which frame in the video sequence is chosen as a starting frame of the video clip for recognition.
As a special case of this process, we can simply perturb the first $P$ frames in every $L$ frames, e.g., 1st to 4th, 33rd to 36th, etc.

Figure \ref{fig:univ_v} shows the results of this attack with respect to the number of videos used to find the universal perturbation.
When 1000 videos are used, the fooling rates exceed 70\% in all cases.
This confirms that the action recognition models are highly vulnerable even in the challenging real-time scenario against the time-invariant universal attack.

It is also worth noting that the attack presented here is more efficient than the existing methods in \cite{li2018adversarial, naeh2020patternless}.
They need to generate perturbations considering all possible cases of temporal offsets, which is not required in the case of the presented attack method.

\section{Conclusion}
\label{sec:conclusion}

We presented in-depth analysis of the structural vulnerability of deep action recognition models against adversarial attack.
Based on the results of perturbing a single frame in a given video clip, we analyzed that the vulnerability is caused by the structural properties such as strides in convolutional layers and max pooling layers and uneven use of inputted frames.
As a result, we demonstrated the possibility of the strong one frame attack that is found to be highly inconspicuous via a subjective experiment.
Finally, we investigated the possibility of finding universal perturbations showing high fooling rates in various attacking scenarios.
\\[-0.4\baselineskip]

\noindent{\large{\textbf{Acknowledgement}}}
\\[-0.6\baselineskip]

This work was supported by the Artificial Intelligence Graduate School Program (Yonsei University, 2020-0-01361).

\clearpage
{\small
\bibliographystyle{ieee_fullname}
\bibliography{main}
}

\end{document}